\documentclass[letterpaper, 10 pt, conference]{ieeeconf}  % Comment this line out if you need a4paper

\IEEEoverridecommandlockouts                              % This command is only needed if 
\title{\LARGE \bf
Trajectory Planning for an Articulated Commercial Vehicle using Model Predictive Contouring Control
}

\author{A.J. Aertssen$^{1,*}$, R.G.M. Huisman$^{1,2}$, I.J.M. Besselink$^{1}$, J. Elfring$^{1}$, and M.J.G. van de Molengraft$^{1}$% <-this % stops a space
\thanks{This work has been carried out within the EU funded Horizon Europe project MODI. Grant agreement ID: 101076810.}% <-this % stops a space
\thanks{$^{1}$Department of Mechanical Engineering, Eindhoven University of Technology, Eindhoven, The Netherlands.}%
\thanks{$^{2}$Safety \& Driver Controls Group, Vehicle Development, DAF Trucks N.V., Eindhoven, The Netherlands.}%
\thanks{$^{*}$Corresponding author: \texttt{a.j.aertssen at tue.nl}}%
}

\usepackage{graphicx}
\usepackage{amsmath}
\usepackage{xcolor}
\usepackage[]{todonotes} % To disable. use [disable]
\usepackage{url}
\usepackage{gensymb}
\usepackage{caption}
\usepackage{subcaption}
\usepackage{lipsum}
\usepackage{bm}
\usepackage{amsfonts}
\usepackage{tikz}
\usepackage{booktabs}
\usepackage{balance}

\newcommand{\Figure}{Fig.~}
\newcommand{\Table}{Table~}
\newcommand{\Sec}{Section~}

\newcommand{\ego}{\textcolor[rgb]{0,0.4470,0.7410}{\rule[0.5ex]{0.5cm}{1.5pt}}}
\newcommand{\obs}{\textcolor[rgb]{0.4660, 0.6740, 0.1880}{\rule[0.5ex]{0.5cm}{1.5pt}}}
\newcommand{\forward}{\textcolor[rgb]{0,0.4470,0.7410}{\rule[0.5ex]{0.5cm}{1.5pt}}}
\newcommand{\reverse}{\textcolor[rgb]{0.8500,0.3250,0.0980}{\rule[0.5ex]{0.5cm}{1.5pt}}}
\newcommand{\qcone}{\textcolor[rgb]{0.4940,0.1840,0.5560}{\rule[0.5ex]{0.5cm}{1.5pt}}}
\newcommand{\qctwo}{\textcolor[rgb]{0,0.4470,0.7410}{\rule[0.5ex]{0.5cm}{1.5pt}}}
\newcommand{\qcthree}{\textcolor[rgb]{0.6350,0.0780,0.1840}{\rule[0.5ex]{0.5cm}{1.5pt}}}

\begin{document}

\maketitle
\thispagestyle{empty}
\pagestyle{empty}

%%%%%%%%%%%%%%%%%%%%%%%%%%%%%%%%%%%%%%%%%%%%%%%%%%%%%%%%%%%%%%%%%%%%%%%%%%%%%%%%
\begin{abstract}

This paper presents a trajectory planning method for articulated commercial vehicles, specifically tractor-semitrailers, based on Model Predictive Contouring Control (MPCC). Although MPCC has proven effective for passenger cars, it is generally ill-suited for tractor-semitrailers. These vehicles are significantly larger, the semitrailer follows a different path than the tractor, and reversing maneuvers are unstable and prone to jackknifing.  Furthermore, practical driving scenarios often require scenario-dependent prioritization of different vehicle `anchor points', e.g., prioritizing the semitrailer position during docking or the tractor position when parking to charge. Therefore, we extend MPCC to enable scenario-dependent weighting of these anchor points and incorporate explicit road-boundary constraints for the front and rear tractor axles and the semitrailer axle, thereby ensuring that all considered wheels remain within the drivable area. The simulation results demonstrate the successful navigation of a representative logistic scenario in both forward and reverse direction. Furthermore, the influence of the optimization parameters on the trajectories is analyzed, providing insights into controlling the vehicle behavior. Finally, first tests using a full-scale prototype vehicle show the practical applicability of the approach.
\end{abstract}

%%%%%%%%%%%%%%%%%%%%%%%%%%%%%%%%%%%%%%%%%%%%%%%%%%%%%%%%%%%%%%%%%%%%%%%%%%%%%%%%
\section{INTRODUCTION} \label{sec:intro}
In logistics, goods are commonly transported using articulated commercial vehicles, such as a tractor-semitrailer. During typical logistics driving scenarios, the vehicle not only drives forward through cities, industrial zones, and highways, but also reverses to load and unload the semitrailer at a dock. The introduction of autonomous vehicles into the logistics chain has the potential to improve efficiency and safety and address driver shortages \cite{Engstrom2019}. A crucial element is a trajectory planner that needs to plan vehicle maneuvers in the aforementioned driving scenarios while avoiding collisions, offering predictable driving to other road users, and maintaining passenger and payload comfort. 
 
\begin{figure}[t]
     \centering
     \begin{subfigure}[b]{0.49\columnwidth}
         \centering
         \includegraphics[width=\linewidth]{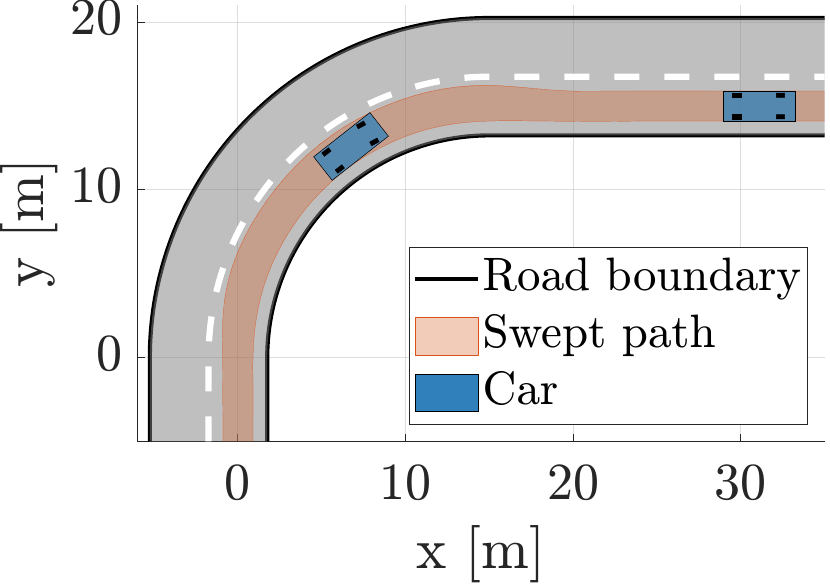}
     \end{subfigure}
     \hfill
     \begin{subfigure}[b]{0.49\columnwidth}
         \centering
         \includegraphics[width=\linewidth]{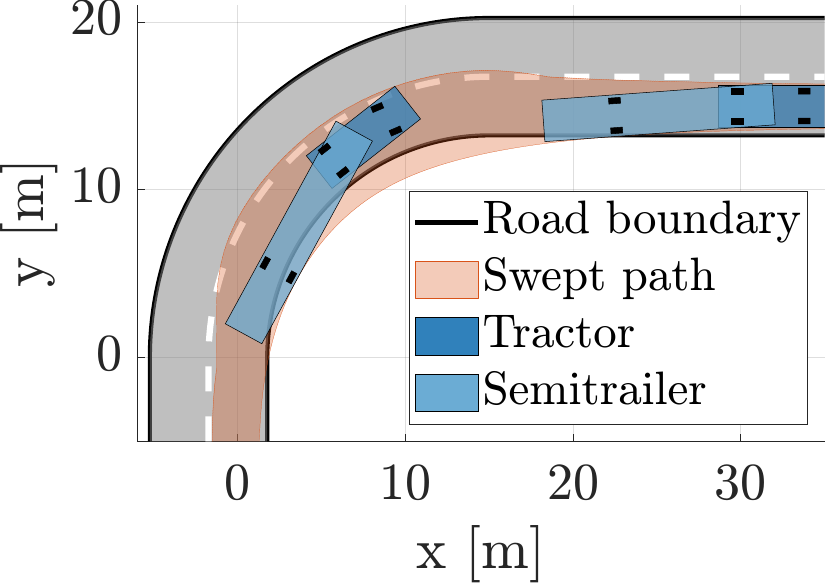}
     \end{subfigure}
    \caption{A car and a tractor-semitrailer navigate a 90\degree~turn with a 15-meter radius on a dual-lane road, where each lane is 3.5 meters wide. The car and tractor follow the center of the right lane with the center of the rear axle. The swept path illustrates the covered area by the vehicles when driving.}
    \label{fig:car_truck}
\end{figure}

Trajectory planning algorithms for passenger cars have been extensively researched \cite{Chen2023, Gonzalez2016, Paden2016, Reda2024}. However, developing a planner for a tractor-semitrailer poses additional challenges. Cars in Europe measure 2.5 to 5.5 meters in length and 1.6 to 2.0 meters in width, while European tractor-semitrailer combinations typically are 16.5 meters long and 2.55 meters wide. Furthermore, the semitrailer follows a different path than the tractor due to the articulation point, which is called offtracking. Moreover, articulated vehicles are unstable while driving in reverse, risking jackknifing \cite{Gosar2023}. The effect of vehicle dimensions and offtracking is illustrated in \Figure \ref{fig:car_truck}, where a car and a tractor-semitrailer negotiate a 90\degree~degree turn on a typical urban street. The car remains in its lane, while the tractor's front extends into the opposite lane, and the semitrailer's inner wheel exits the road on the inside. The example illustrates the need to explicitly consider the dimensions of the vehicle and the offtracking of the semitrailer during trajectory planning. 

Any autonomous vehicle should behave in a predictable way. Other road users expect that an autonomous vehicle generally follows the middle of the lane at approximately the speed limit. We assume that a local geometric road model is available, which is a simplified, high-resolution map of the road geometry (e.g., lanes) including speed limits within the vehicle's immediate vicinity. We rely on HD-maps to create the road model but it could also be constructed from onboard sensor data (e.g., LiDAR, camera) and vehicle-to-everything (V2X) communication. A trajectory planner can leverage the information from the local road model to plan a trajectory that the tractor-semitrailer must follow and match the expectations of the other road users. The \textit{trajectory} should be:
\begin{enumerate}
    \item safe, i.e. avoid collisions with static and dynamic obstacles and prevent rollover of the vehicle,
    \item comfortable for the passengers and payload,
    \item efficient in terms of travel time.
\end{enumerate} 
These requirements can conflict. For example, a longer travel time may result from the vehicle slowing down to prevent rollover while cornering. Furthermore, the planner must generate trajectories that are kinematically and dynamically feasible for a tractor-semitrailer for both forward and backward motion (e.g., no instantaneous 90\degree~turns).

A \textit{trajectory planner} for an articulated commercial vehicle should be able to prioritize one or more vehicle anchor points based on the driving scenario, where vehicle anchor points are specific physical locations on the vehicle. For example, the semitrailer’s rear-door position is most important during docking, the tractor’s charging-port location is crucial when charging, and both the tractor and the semitrailer should remain close to the lane center during highway driving. 

In this paper, we propose a trajectory planner that meets these requirements to plan a safe, comfortable, and efficient trajectory. The remainder of this paper is structured as follows. First, related work is discussed in \Sec \ref{sec:RW} and \Sec \ref{sec:prelim} presents the necessary preliminaries. The approach itself is formulated in \Sec \ref{sec:methods}, followed by its application to a common logistics scenario in simulation and a first real-life test in \Sec \ref{sec:results}. Finally, conclusions and future work are provided in \Sec \ref{sec:conc}.

\section{RELATED WORK} \label{sec:RW}
A variety of vehicle trajectory planning approaches have been developed, including graph search, sampling-based methods, interpolation, and optimization \cite{Chen2023, Gonzalez2016, Paden2016, Reda2024}. As noted in the introduction, a trajectory planner for tractor-semitrailers requires handling potentially conflicting objectives while ensuring that constraints related to safety and comfort are met. This renders graph-based, sampling, and interpolation methods unsuitable, while optimization-based motion planning techniques are particularly suitable for these requirements \cite{Chen2023, Gonzalez2016, Paden2016, Reda2024}. These methods can also ensure the kinematic and dynamic feasibility of the trajectory through explicit use of vehicle model constraints.

\begin{figure}[t]
    \centering
    \begin{subfigure}{0.48\columnwidth}
        \centering
        \includegraphics[width=\linewidth]{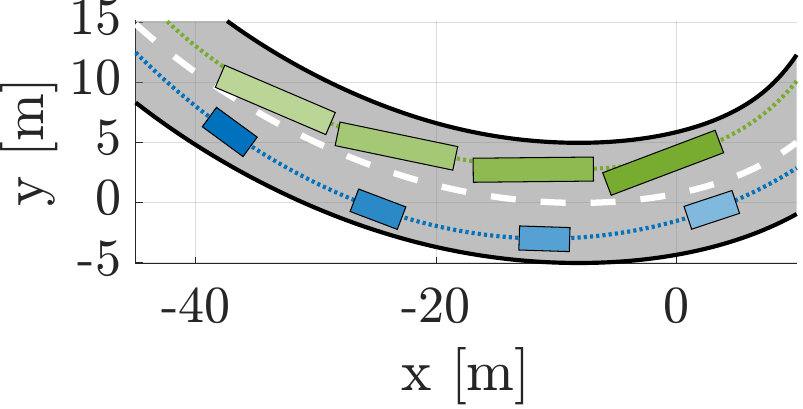}
    \end{subfigure}
    \hfill
    \begin{subfigure}{0.48\columnwidth}
        \centering
        \includegraphics[width=\linewidth]{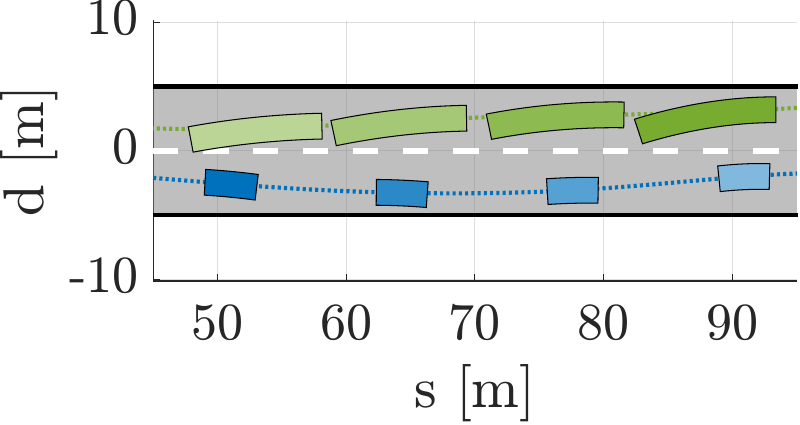}
    \end{subfigure}

    \caption{Driving scenario with a car (\ego) and a longer vehicle (\obs) on a road with varying curvature, illustrated in Cartesian (left) and Frenet (right) frames. Due to the nonlinear transformation between the frames, obstacles that are convex in Cartesian coordinates may become non-convex in Frenet coordinates, with their shapes depending on the local path curvature.}
    \label{fig:example_coordinates}
\end{figure}

Optimization-based methods developed for cars, such as \cite{Laurense2022, Micheli2023}, are generally not suitable for large articulated vehicles due to their size, offtracking, and jackknifing. Existing optimization-based approaches for articulated vehicles typically use either Cartesian or road-aligned (Frenet) coordinate frames, both illustrated in \Figure \ref{fig:example_coordinates}. The Cartesian frame is used by Bos et al. \cite{Bos2023} in a Model Predictive Control (MPC) approach, where the free space is decomposed into convex polyhedrons to constrain the tractor and semitrailer positions. The Cartesian frame enables straightforward obstacle representation using convex shapes, facilitating efficient obstacle avoidance constraints in an optimization problem. In contrast, the road-aligned frame allows a simple definition of road boundaries and reference paths, since these curves are typically aligned with the road or lane center \cite{Reiter2023}. This frame is employed in Duijkeren et al. \cite{Duijkeren2015}, where nonlinear MPC is applied for highway scenarios. However, their approach is limited to low-curvature environments due to underlying assumptions. Oliveira et al. \cite{Oliveira2020} also use a road-aligned frame for scenarios involving narrow passages and sharp turns, focusing solely on forward driving. Although the road-aligned frame facilitates the straightforward integration of the information from the local road model, it requires complex approximations of the position of the semitrailer axle. Additionally, it complicates obstacle avoidance constraints, which potentially introduces conservative constraints and restricts feasible maneuvers \cite{Reiter2023}. 

We leverage the advantages of both coordinate frames by using Model Predictive Contouring Control (MPCC) \cite{Lam2010}. MPCC integrates both coordinate frames by employing a vehicle prediction model in Cartesian coordinates, augmented with an additional state that represents the transformation between the two coordinate systems. This enables obstacle avoidance constraints to be expressed in Cartesian coordinates and road boundary constraints in Frenet coordinates. Within the domain of vehicle trajectory planning, MPCC has so far only been applied to passenger cars \cite{Bertipaglia2024, Ferranti2019, Liniger2014, Pauls2022}. Most of these approaches simplify the vehicle to a single point for boundary constraints \cite{Bertipaglia2024, Ferranti2019, Liniger2014}, although Pauls et al. \cite{Pauls2022} constrain multiple vehicle anchor points. Because their method assumes a car that stays close to a reference path, it cannot be applied to articulated commercial vehicles. Moreover, existing MPCC methods cannot ensure that the wheels of articulated vehicles remain within the road boundaries.

Therefore, in this study, we extend the existing MPCC framework by introducing additional augmented states. This enables the explicit formulation of road boundary constraints for the front and rear tractor axles, as well as the semitrailer axle, ensuring that all considered wheels remain within the drivable space. Furthermore, this extension allows to prioritize vehicle anchor points based on the driving scenario and prevent jackknifing.

\section{PRELIMINARIES} \label{sec:prelim}
\begin{figure}[t]
    \centering
    \includegraphics[width=0.8\columnwidth]{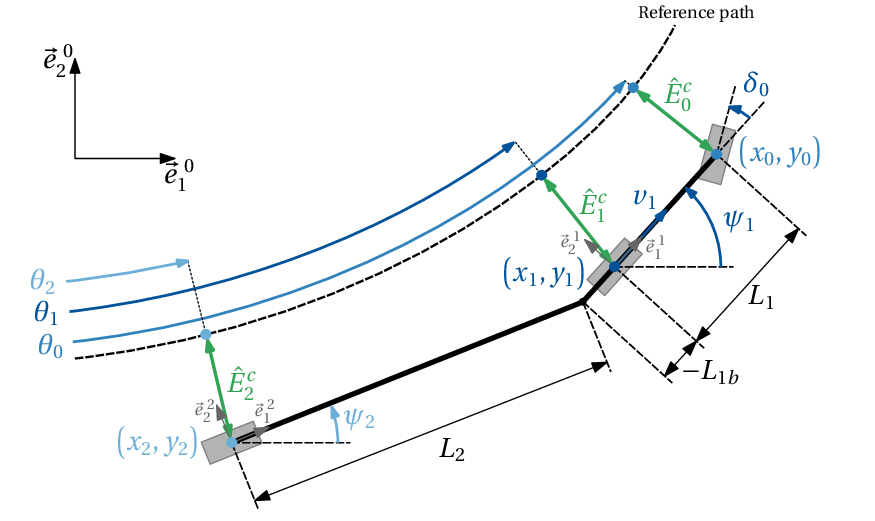}
    \caption{Visualization of the kinematic tractor-semitrailer vehicle model, where $L_1$ is the wheelbase of the tractor measured in $\vec{e}^{\,1}$, $L_{1b}$ is the articulation joint position w.r.t. the rear axle in $\vec{e}^{\,1}$, and $L_2$ is the wheelbase of the semitrailer measured in $\vec{e}^{\,2}$. Additionally, the augmented states $\theta_i$ and the corresponding contour errors $\hat{E}^{c}_{i}$ for $i\in\{0,1,2\}$ are shown. In the illustration, the articulation joint is positioned behind the rear axle for clarity, while, in practice, it is commonly positioned in front of the rear axle.}
    \label{fig:MPCC_multiple_points}
\end{figure}

We denote vectors in \textbf{bold}. The global Cartesian coordinate frame is denoted by $\vec{e}^{\,0}$, the body-fixed coordinate frames of the tractor and the semitrailer are $\vec{e}^{\,1}$ and $\vec{e}^{\,2}$, respectively, as illustrated in \Figure \ref{fig:MPCC_multiple_points}.

The subscript $k\in \mathbb{N}$ denotes the discrete timestep, where each timestep corresponds to a discrete time instance $ t_k = k \Delta t $, with $\Delta t$ being a fixed sampling time. The subscript $i$ refers to the index of the anchor points of the vehicle, where $i \in \{0,1,2\}$ corresponds to the front axle of the tractor ($i=0$), the rear axle of the tractor ($i=1$) and the semitrailer axle ($i=2$), as depicted in \Figure \ref{fig:MPCC_multiple_points}.

We assume that an upstream module in the autonomous driving pipeline generates a drivable corridor using the local road model. This corridor is defined by a reference path, representing the path the vehicle is expected to follow, and by the lateral distances from this path to the left and right boundaries. These boundaries indicate the area within which the vehicle's wheels should remain and can be based on the ego lane edges, road boundaries, or other relevant boundaries. The reference path is parameterized by its arc length $s$ using spline parametrization, which allows obtaining any point $\left(x_r(s), y_r(s)\right)$ on the reference path as a function of $s$. The angle $\psi_r(s)$ of the tangent to the path is given by
\begin{equation}
    \psi_r(s) = \arctan\left( \frac{y'_r(s)}{ x'_r(s)}\right),
\end{equation}
where $y'_r(s)$ and $x'_r(s)$ are the derivatives of $y_r$ and $x_r$ with respect to $s$, respectively \cite{Schwarting2018}. Similarly, it is possible to obtain the perpendicular lateral distance from the reference path to the left boundary $b_l(s)$ and the right boundary $b_r(s)$.

A kinematic bicycle model for the tractor-semitrailer is used, as illustrated in \Figure \ref{fig:MPCC_multiple_points}. Here, the position of the tractor rear axle is given by $x_1\vec{e}^{\,0}_1 + y_1\vec{e}^{\,0}_2$, with $x_1$ and $y_1$ being the respective components along $\vec{e}^{\,0}_1$ and $\vec{e}^{\,0}_2$. The heading angles of the tractor and the semitrailer are denoted by $\psi_1$ and $\psi_2$, respectively. The longitudinal velocity of the tractor is represented by $v_1$, and $\delta_0$ denotes the steering angle of the tractor's front wheel. The state and input vectors are $\bm{x} =\left[ x_{1},y_{1},\psi_{1},\psi_{2}\right]^\top$ and $\bm{u} =\left[v_{1}, \delta_0 \right]^\top$, the continuous time kinematic equations are given by
\begin{equation} \label{eq:veh_model}
    \begin{bmatrix}
        \dot{x}_1 \\ \dot{y}_1 \\ \dot{\psi}_1 \\ \dot{\psi}_2
    \end{bmatrix}
   = \begin{bmatrix}
        v_1 \cos{(\psi_1)} \\
        v_1 \sin{(\psi_1)} \\
        \frac{v_1}{L_1} \tan{(\delta_0)} \\
        \frac{v_1}{L_2} \sin(\gamma_1) + \frac{L_{1b}}{L_2}\dot{\psi}_1\cos{(\gamma_1)}
    \end{bmatrix},
\end{equation}
where $\gamma_1 = \psi_1-\psi_2$ is the articulation angle, and $L_1,~L_{1b}$ and $L_2$ represent vehicle dimensions as defined in \Figure \ref{fig:MPCC_multiple_points}. Note that we assume that the three (non-steered) semitrailer axles can be approximated by a single axle.

\section{PLANNING PROBLEM FORMULATION} \label{sec:methods}
The planning problem is formulated as a constrained nonlinear optimization problem, where an objective function \eqref{eq:cost} is minimized while being subject to vehicle model dynamics \eqref{eq:con_veh}, inequality constraints concerning road boundaries, comfort, and safety \eqref{eq:ineq_con}, and state and input bounds \eqref{eq:state_con}-\eqref{eq:input_con}:
\begin{subequations}
    \begin{align}
        \min_{U_k} \quad & J\left(X_k ,U_k\right),  &  \label{eq:cost}\\
        \text{s.t.} \quad & \bm{x}^{m}_{j+1|k} = f \left(\bm{x}^{m}_{j|k},\bm{u}^{m}_{j|k}\right) & j &= 0,...,N-1, \label{eq:con_veh}\\
        & \bm{x}^{m}_{0|k} = \bm{x}^m_{k}, & \label{eq:con_init} \\
        & \bm{g}\left(\bm{x}^{m}_{j|k},\bm{u}^{m}_{j|k}\right) \leq 0 & j &= 0,...,N, \label{eq:ineq_con}\\
        & \underline{\bm{x}} \leq \bm{x}^{m}_{j|k} \leq \overline{\bm{x}} & j &= 0,...,N, \label{eq:state_con} \\
        & \underline{\bm{u}} \leq \bm{u}^{m}_{j|k} \leq \overline{\bm{u}} & j &= 0,...,N-1. \label{eq:input_con}
    \end{align}
\end{subequations}
Here, $N \in \mathbb{N}_{\geq 1}$ is the prediction horizon. The states and inputs at time $k+j$ are denoted by $\bm{x}^{m}_{j|k}$ and $\bm{u}^{m}_{j|k}$, where the superscript $m$ refers to the states and inputs in the MPCC formulation, which will be defined later. $X_k$ is the sequence of predicted states defined as $X_k = \left\{\bm{x}_{1|k}, ..., \bm{x}_{N|k} \right\}$, which depends on the initial condition $\bm{x}^m_{0|k}$, the discretized vehicle kinematics represented by $f(\cdot)$, and the predicted input sequence defined as $U_k = \left\{\bm{u}_{0|k}, ..., \bm{u}_{N-1|k} \right\}$. The objective function is denoted by $J(\cdot)$. The inequality constraints are collectively represented by $g(\cdot)$, and the lower and upper bounds on the states and inputs are denoted by $\underline{\bm{x}}, \overline{\bm{x}}, \underline{\bm{u}},$ and $\overline{\bm{u}}$, respectively.

\begin{figure}[t]
    \centering
    \includegraphics[width=0.8\columnwidth]{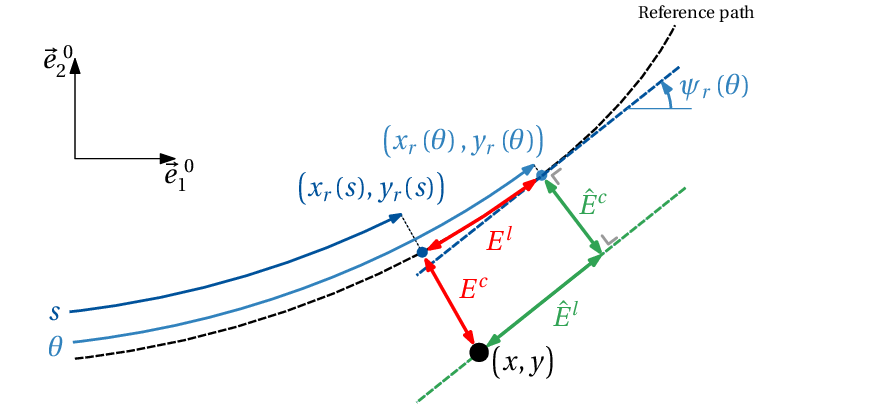}
    \caption{Conceptual sketch of the MPCC approach for an arbitrary point $(x,~y)$, where the dashed line represents the reference path. The red arrows are the contour and lag errors in the Frenet frame and the green arrows are the approximation of those errors used in the MPCC formulation.}
    \label{fig:MPCC_error}
\end{figure}

In general, the goal of MPCC is to steer a point $\left(x, y \right)$ along a reference path, as visualized in \Figure \ref{fig:MPCC_error}. This is achieved by minimizing the contour error $E^{c}_{k}$ and the lag error $E^{l}_{k}$, while maximizing the path speed \cite{Lam2010}. The errors are a function of $s$, which is the traveled distance by $\left(x, y \right)$ along the reference path. The calculation of $s$ requires the orthogonal projection of $\left(x, y \right)$ onto the reference path, which would lead to a nested optimization problem. The resulting computational load is typically too high to be used in dynamic, fast-changing contexts like city driving \cite{Lam2010, Schwarting2018}. Therefore, in MPCC, the state vector $\bm{x}$ is augmented with the progress state $\theta$. The dynamics of $\theta$ are given by
\begin{equation}
    \dot{\theta} = v^\theta,
\end{equation}
where the velocity of the progress state $v^\theta$ is added to the input vector $\bm{u}$. By adjusting the virtual velocity $v^\theta$, it is ensured that $E_k^l$ is minimized such that the progress variable $\theta$ approximates $s$. This allows to approximate the lag and contour error by $\hat{E}^{l}$ and $\hat{E}^{c}$ \cite{Pauls2022}:
\begin{subequations}\label{eq:errors}
    \begin{align}
        \begin{split}
            \hat{E}^{c} = -&\left(x - x_r\left(\theta\right)\right)\sin\left(\psi_r\left(\theta\right)\right) +\\ 
            \quad&\left(y - y_r\left(\theta\right)\right)\cos\left(\psi_r\left(\theta\right)\right),
        \end{split}\\
        \begin{split}
            \hat{E}^{l} = &\left(x - x_r\left(\theta\right)\right)\cos\left(\psi_r\left(\theta\right)\right) +\\ 
            \quad &\left(y - y_r\left(\theta\right)\right)\sin\left(\psi_r\left(\theta\right)\right).
        \end{split}
    \end{align}
\end{subequations}
From \Figure \ref{fig:MPCC_error}, it can be observed that $\theta \approx s$ if $\hat{E}^{l} \approx 0$. In contrast, $\hat{E}^c$ must be allowed to be comparably large to, for example, avoid obstacles on the reference path. These two conditions will be enforced in the objective function, which will be discussed in Section \ref{ssec:obj}.

\subsection{Equality Constraints}
The equality constraints consider the vehicle model. We extend the vehicle model in \eqref{eq:veh_model} with integrators for inputs $v_1$ and $\delta_0$ to be able to impose limits on their rate of change, which is a common approach in Model Predictive Control \cite{Rawlings2017}. This ensures that the planned trajectory can actually be executed by a tractor‑semitrailer, given its limits on steering speed and acceleration capabilities. Therefore, the vehicle model in \eqref{eq:veh_model} is extended with 
\begin{equation}
\begin{split}
 \dot{v}_1 &= a_1, \\
 \dot{\delta}_0 &= \dot{\delta}^u_0,
\end{split}
\end{equation}
where $a_1$ is longitudinal acceleration of the tractor and $\dot{\delta}^u_0$ is the steering rate. The model is discretized using a multiple shooting approach with fourth-order Runge-Kutta integration \cite{LaValle2006}. Consequently, the vehicle model is formulated as a general discrete dynamical system as given in \eqref{eq:con_veh}.

\begin{figure}[t]
    \centering
    \includegraphics[width=0.8\columnwidth]{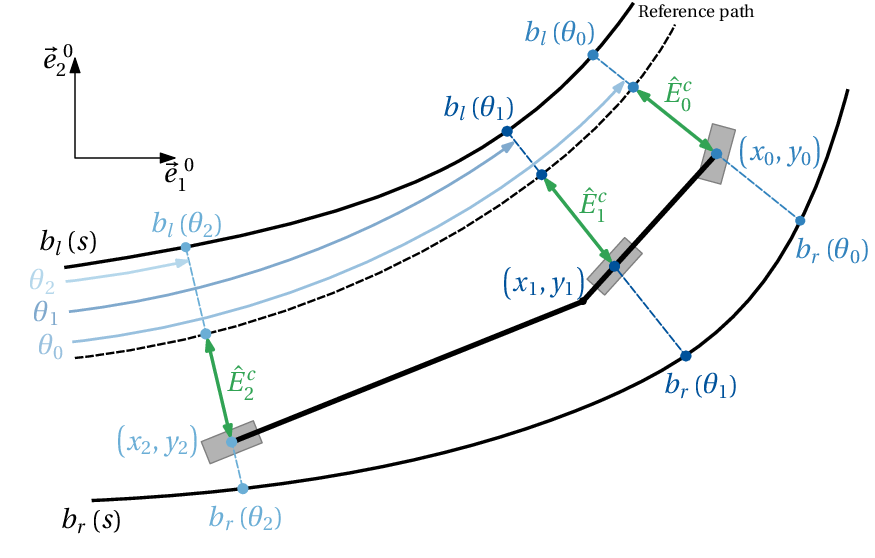}
    \caption{Visualization of the corridor boundary constraints for each vehicle anchor point $i$.}
    \label{fig:MPCC_boundaries}
\end{figure} 

\subsection{Objective Function} \label{ssec:obj}
The objective function in \eqref{eq:cost} is given by
\begin{equation} \label{eq:total_obj_fun}
\begin{split}
        J\left(X_k ,U_k\right) &= J^{m} \left(X_k,U_k\right) + J^{c} \left(U_k\right),
\end{split}
\end{equation}
where $J^{m}$ represents the MPCC objective function and $J^{c}$ is the objective function representing passenger and payload comfort. We do not explicitly include energy consumption, despite its importance to truck manufacturers and logistics organizations, but it can be added to the cost function if desired.

The objective function $J^{m}_{i}$ for one vehicle anchor point $i$ quadratically weights the lag and contour error, combined by a linear progress maximization reward to maximize path speed \cite{Pauls2022}:
\begin{equation} \label{eq:error_i}
    J^{m}_{i}\left(X_k,U_k\right) = \sum_{j=1}^{N} \bm{e}^\top_{i,j} Q_i \bm{e}_{i,j} - \sum_{j=0}^{N-1} q^{v}_{i} v^{\theta}_{i,j},
\end{equation}
where $\bm{e}_{i,j} = [\hat{E}^{c}_{i}(\bm{x}^{m}_{j|k}), \hat{E}^{l}_{i}(\bm{x}^{m}_{j|k})]^\top$ and $\hat{E}^{c}_{i}$ and $\hat{E}^{l}_{i}$ are defined in \eqref{eq:errors}. $Q_i = \text{diag}(q^{c}_{i}, q^{l}_{i})$, which is a diagonal matrix with entries $q^{c}_{i}$ and $q^{l}_{i}$ on the diagonal and zeros elsewhere. The weights $q^{c}_{i}$, $q^{l}_{i}$, and $q^{v}_{i}$ weigh the contour error, the lag error, and the progress velocity, respectively, for vehicle anchor point $i$. The separate lag and contour weights ensure that $\hat{E}^{l}_i$ remains sufficiently small such that $\theta_i \approx s$, while allowing $\hat{E}^{c}_{i}$ to be large enough to leave the reference path if necessary.

In our research, we consider three vehicle anchor points $i\in\{0,1,2\}$, which are visualized in \Figure \ref{fig:MPCC_multiple_points}. As a result, there are three augmented states $\theta_i$ with corresponding velocity $v^{\theta}_{i}$, so we define $\bm{x}^{m}=\left[ x_{1},y_{1},\psi_{1},\psi_{2},v_{1},\delta_0, \theta_0, \theta_1, \theta_2 \right]^\top$ and $\bm{u}^{m} = [a_1,\dot{\delta}^u_0,v^{\theta}_{0}, v^{\theta}_{1}, v^{\theta}_{2}]^\top$. The MPCC objective function in \eqref{eq:total_obj_fun} is then equal to 
\begin{equation} 
    J^{m}\left(X_k,U_k\right) = \sum_{i=0}^{2} J^{m}_{i}\left(X_k,U_k\right).
\end{equation}
Each anchor point is associated by a dedicated set of weights, allowing prioritizing different parts of the vehicle, such as the semitrailer during docking maneuvers or both the tractor and the semitrailer during highway driving. By assigning driving scenario‑specific weights, this general approach can realize distinct behaviors (e.g., prioritizing the semitrailer during docking) across diverse driving scenarios without the need to develop specific methods for each scenario.

The comfort term $J^{c}$ quadratically penalizes large acceleration and steering rate values as a measure of passenger and payload comfort:
\begin{equation}
        J^{c}\left(U_k \right) = \sum_{j=0}^{N-1} 
 \left(\bm{u}^{m}_{j|k}\right)^\top R \bm{u}^{m}_{j|k},
\end{equation}
where $R = \text{diag}(q^a, q^{\dot{\delta}}, 0, 0, 0)$ is a weight matrix. 

\subsection{Inequality Constraints}
The states and control inputs have to adhere to the limits given in \eqref{eq:state_con} and \eqref{eq:input_con}. The limit values are determined considering comfort, actuator constraints, and legislation:
\begin{equation}
    \begin{split}
        &\left|\delta_{0} \right| \leq \delta_{0,\text{max}}, \\
        &\left|\dot{\delta}^u_{0} \right| \leq \dot{\delta}^u_{0,\text{max}}, \\
        &v_{1,\text{min}} \leq v_{1} \leq v_{1,\text{max}}, \\
        &\left|a_{1} \right| \leq a_{1,\text{max}}, \\
        %&0 \leq \theta_i \leq \theta_{i, \text{max}},~\forall i\in \{0,1,2\}, \\
        %&0 \leq v^\theta_i \leq v^\theta_{i, \text{max}},~\forall i\in \{0,1,2\}.
    \end{split}
\end{equation}

To keep the vehicle within the given corridor, the contour error $\hat{E}^c_{i}$ is constrained using $b_l(\theta_i)$ and $b_r(\theta_i)$ in \eqref{eq:ineq_con}, as visualized in \Figure \ref{fig:MPCC_boundaries}. This constraint is denoted for vehicle anchor point $i$ as
\begin{equation} \label{eq:corridor_cons}
    b_r(\theta_{i}) + \frac{w}{2} \leq \hat{E}^c_{i} \leq b_l(\theta_{i}) - \frac{w}{2},
\end{equation}
where $w$ is the vehicle width. We assume that the tractor and the semitrailer have the same width that remains constant along the vehicle. The constraints ensure that considered axle positions remain within boundaries, while overhanging sections may extend beyond. The left and right boundaries are assumed to account for vehicle overswaying. 

To enforce passenger comfort and to ensure that the vehicle does not tip over, the lateral acceleration is constrained in \eqref{eq:ineq_con}:
\begin{equation}
    \left|v_{1}\dot{\psi}_{1}\right| \leq a_{y,\text{max}},
\end{equation}
where $a_{y,\text{max}}$ is the maximum absolute lateral acceleration.

\begin{table}[t]
\centering
\caption{Optimization problem parameters. The velocity limits are given in \texttt{forward} $\mid$ \texttt{reverse} driving.}
\label{tab:MPCCparam}
\begin{tabular}{p{0.2\columnwidth} p{0.3\columnwidth} p{0.08\columnwidth} p{0.22\columnwidth}}
\toprule
\multicolumn{2}{c}{\textbf{Vehicle}} & \multicolumn{2}{c}{\textbf{Optimization}} \\
\midrule
$L_1$, $L_{1b}$, $L_{2}$ & 4.0~m, 0.6~m, 8.0~m & $N$ & 75 \\
$w$ & 2.5~m & $T$ & 15~s \\
$v_{1,\text{min}}$ & 0 $\mid$ -5~km~h$^{-1}$ & $\Delta t$ & 0.2~s \\
$v_{1,\text{max}}$ & 15 $\mid$ 0~km~h$^{-1}$ & $q^l_0,q^c_0$ & 5000,~5 \\
$a_{1,\text{max}}$ & 0.5~m~s$^{-2}$ & $q^l_1,q^c_1$ & 5000,~5 \\
$\delta_\text{max}$ & 40~\degree & $q^l_2,q^c_2$ & 5000,~100 \\
$\dot{\delta}^u_\text{max}$ & 20~\degree~s$^{-1}$ & $R$ & diag(1,10,0,0,0) \\
$a_{y,\text{max}}$ & 1.5~m~s$^{-2}$ & $q_i^v$ & 1 \\
\bottomrule
\end{tabular}
\end{table}

\begin{figure}[t]
\centering
\begin{minipage}[t]{0.57\columnwidth}
    \vspace{0pt}%
    \includegraphics[width=\linewidth]{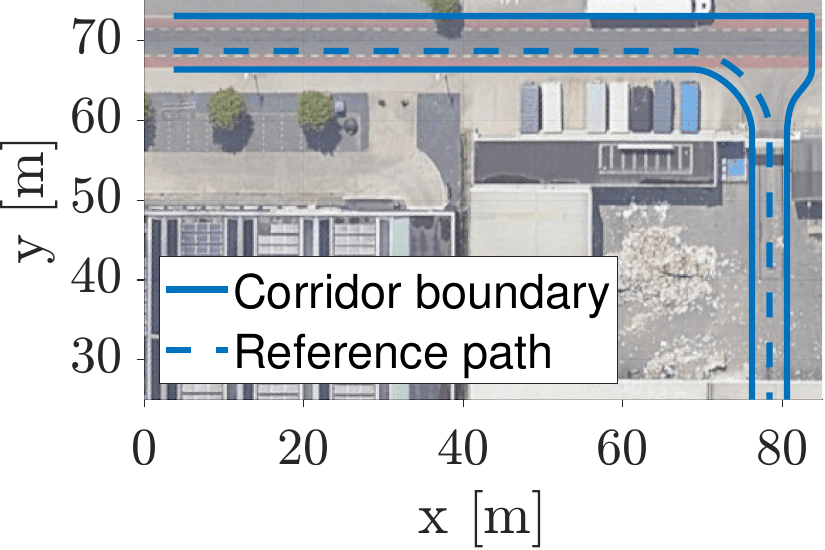}
    \caption{Visualization of the common logistics driving scenario, featuring a manually defined drivable corridor.}
    \label{fig:corridor}
\end{minipage}
\hfill
\begin{minipage}[t]{0.40\columnwidth}
    \vspace{0pt}%
    \includegraphics[width=\linewidth]{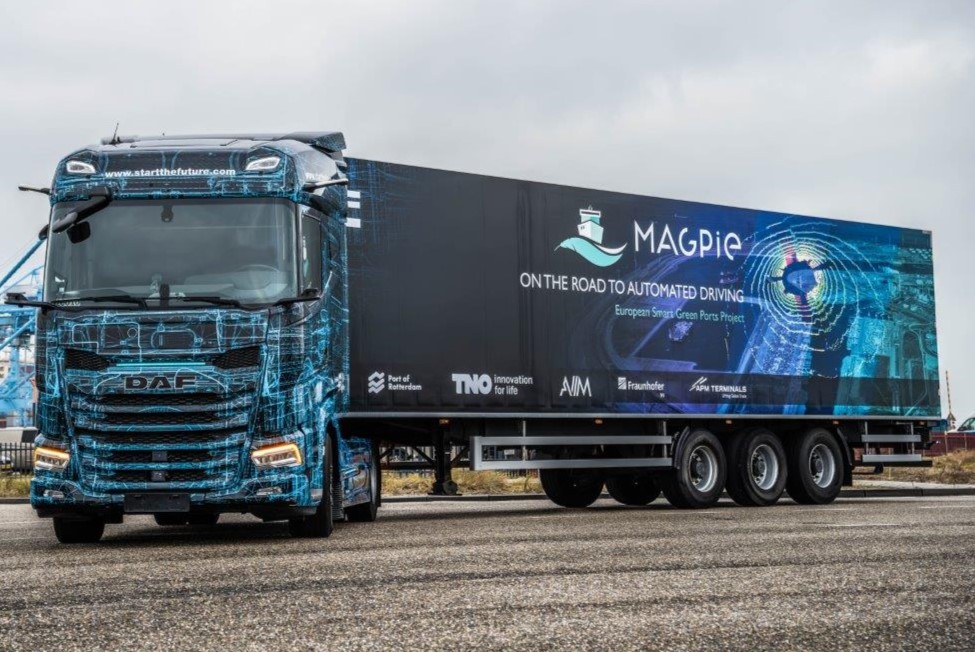}
    \vspace{6mm}
    \caption{Prototype vehicle (© DAF Trucks N.V.).}
    \label{fig:proto}
\end{minipage}
\end{figure}

\section{Results} \label{sec:results}
\begin{figure}[t]
    \centering
    \begin{subfigure}{0.48\columnwidth}
        \centering
        \includegraphics[width=0.95\linewidth]{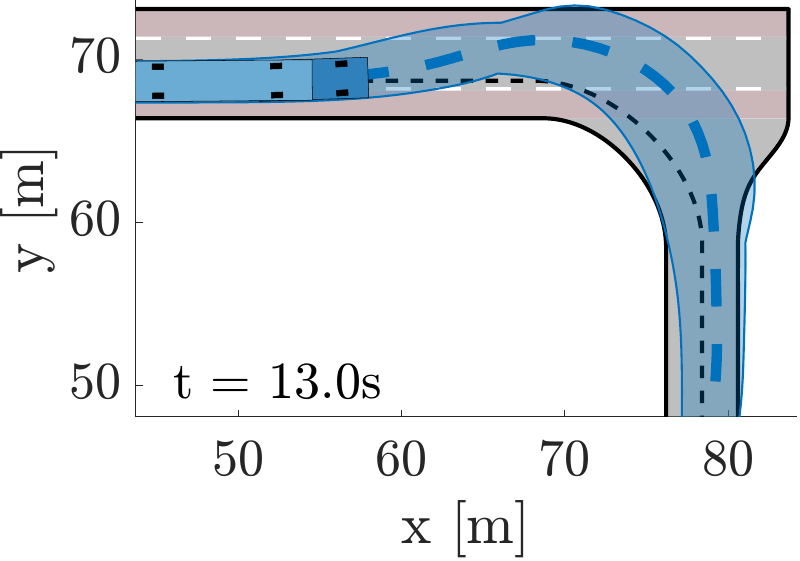}
    \end{subfigure}
    \hfill
    \begin{subfigure}{0.48\columnwidth}
        \centering
        \includegraphics[width=0.95\linewidth]{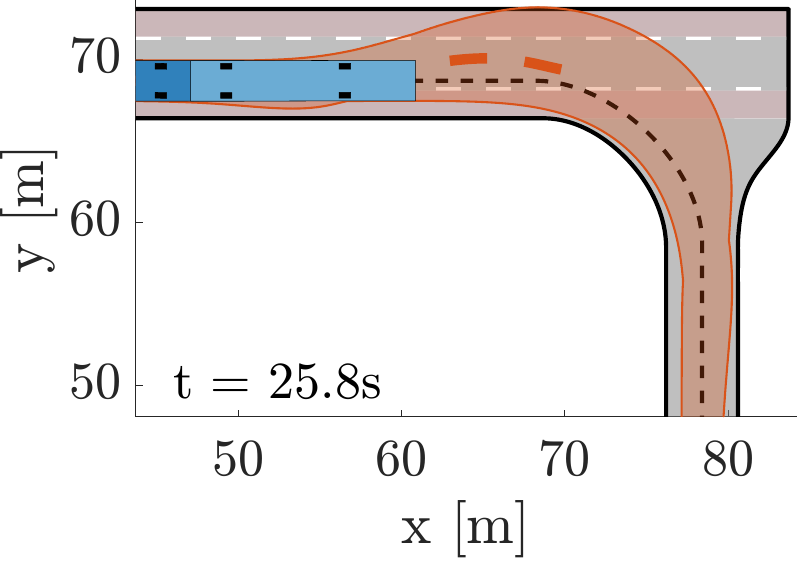}
    \end{subfigure}

    \vspace{0.5em}

    \begin{subfigure}{0.48\columnwidth}
        \centering
        \includegraphics[width=0.95\linewidth]{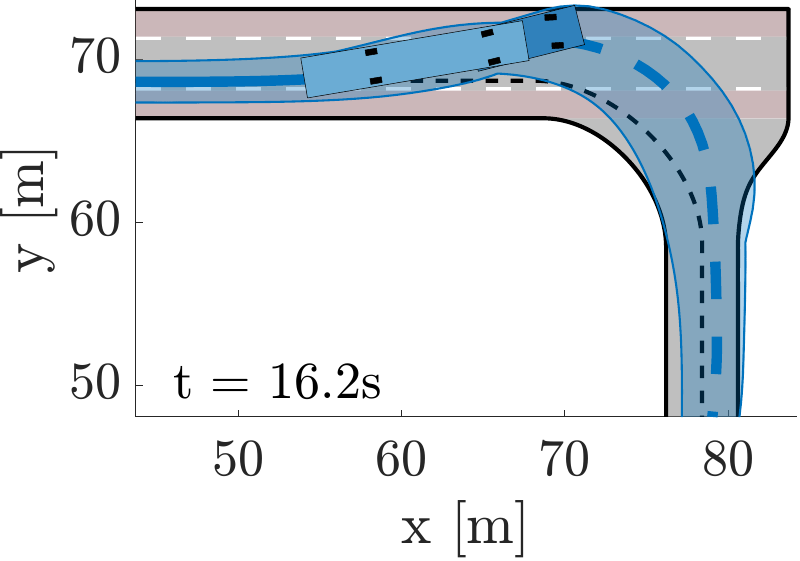}
    \end{subfigure}
    \hfill
    \begin{subfigure}{0.48\columnwidth}
        \centering
        \includegraphics[width=0.95\linewidth]{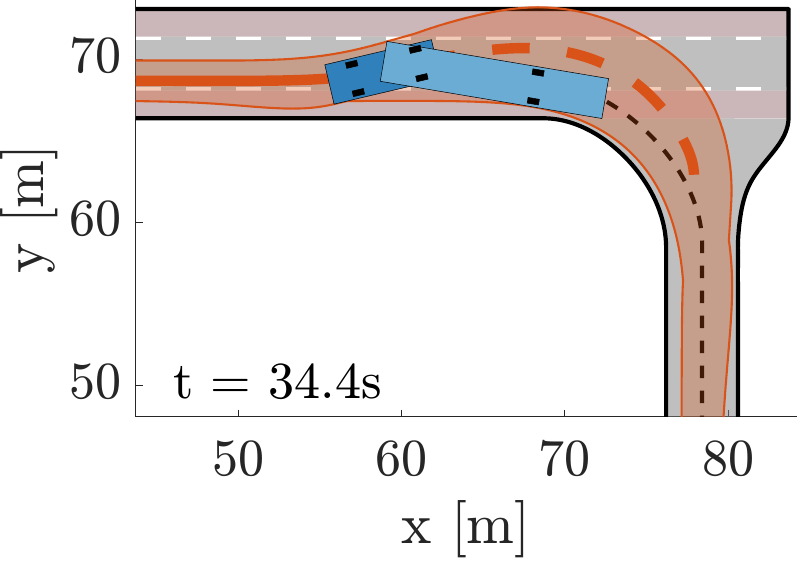}
    \end{subfigure}

    \vspace{0.5em}

    \begin{subfigure}{0.48\columnwidth}
        \centering
        \includegraphics[width=0.95\linewidth]{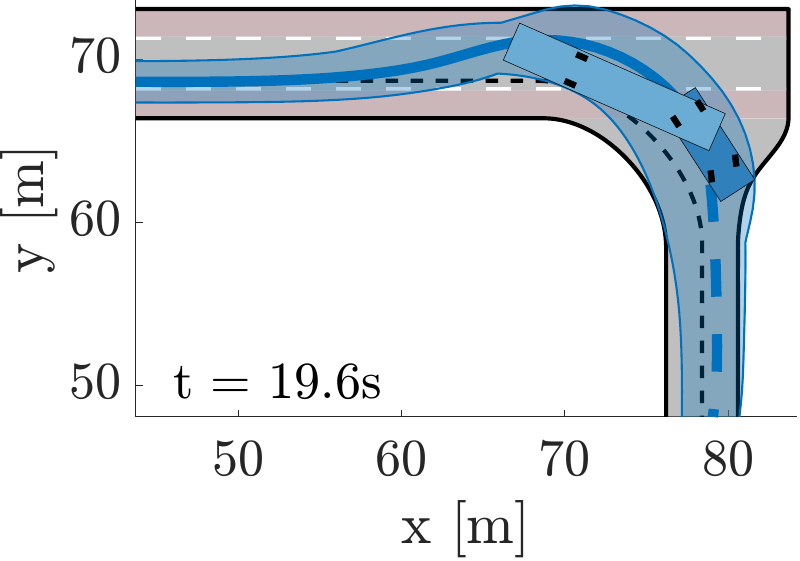}
    \end{subfigure}
    \hfill
    \begin{subfigure}{0.48\columnwidth}
        \centering
        \includegraphics[width=0.95\linewidth]{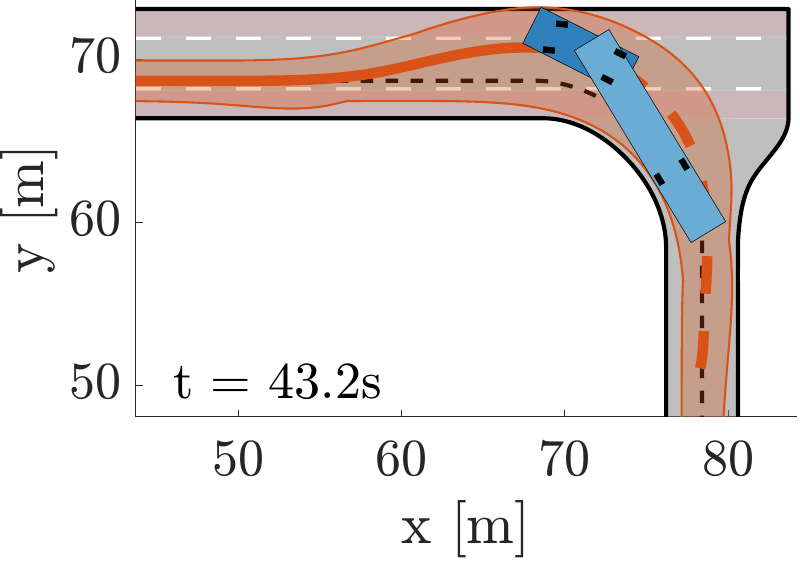}
    \end{subfigure}

    \vspace{0.5em}

    \begin{subfigure}{0.48\columnwidth}
        \centering
        \includegraphics[width=0.95\linewidth]{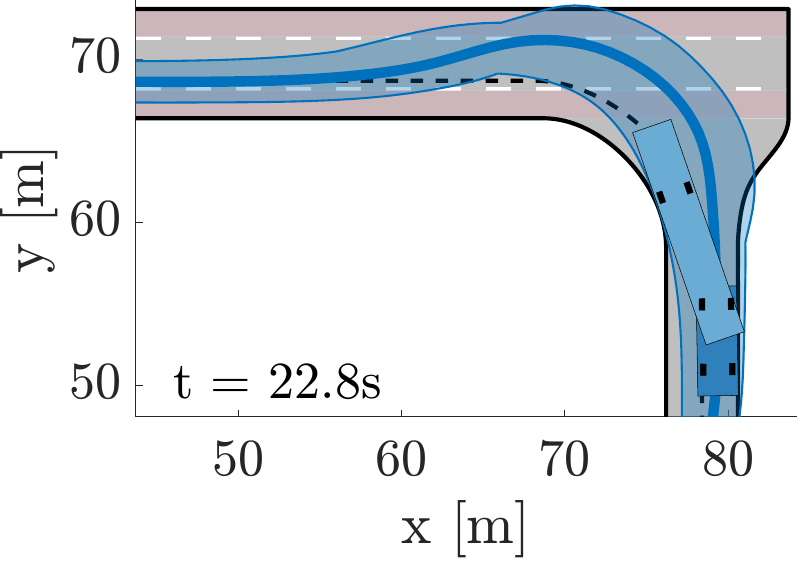}
    \end{subfigure}
    \hfill
    \begin{subfigure}{0.48\columnwidth}
        \centering
        \includegraphics[width=0.95\linewidth]{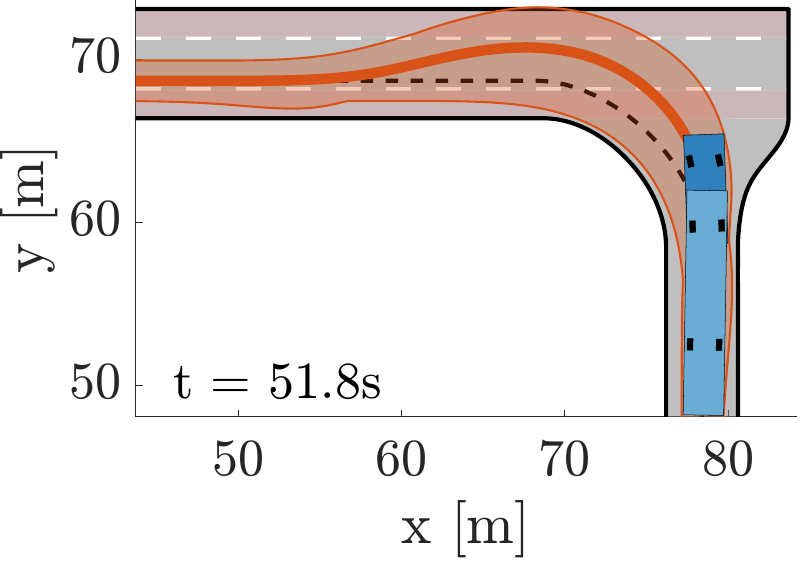}
    \end{subfigure}

    \caption{Simulation results for forward driving (\forward) and reverse driving (\reverse) scenarios, showing the driven (solid) and predicted (dashed) paths of the rear axle over time. The reference path is plotted as a black dashed line, and the shaded area visualizes the swept path of the full vehicle body during the maneuver. Because the reserve maneuver is infeasible starting from the proper lane, the reverse scenario starts from the opposite side of the road.}
    \label{fig:results_DD}
\end{figure}

The proposed MPCC algorithm is implemented in Python using a receding horizon scheme, where the optimal control problem is repeatedly solved over a moving time window using CasADi \cite{Andersson2019} and the \texttt{HSL\_MA57} solver \cite{HSL2025} in IPOPT \cite{Wächter2006}. The relevant parameters are given in \Table \ref{tab:MPCCparam}. The prediction time $T$ is a tradeoff between computational complexity and sufficient prediction distance in both driving directions. The computation time is in the order of 100~ms per iteration (Intel i7-11800H CPU, Ubuntu 20.04). The horizon time combined with the maximum velocity of 15~km/h provides the trajectory planner with a forward look-ahead distance equivalent to approximately three tractor-semitrailer lengths. At a reverse-driving speed limit of 5 km/h, the look-ahead distance corresponds to approximately one tractor-semitrailer length. 

We apply the MPCC method in simulation to a common logistics scenario illustrated in \Figure \ref{fig:corridor}, where a vehicle must enter the premises of a customer through a gate or an industrial roll-up door from a city road. \Figure \ref{fig:corridor} also shows the drivable corridor. This scenario lets us demonstrate that the proposed MPCC algorithm plans a safe, comfortable, and efficient trajectory by ensuring the vehicle’s wheels stay within the corridor boundaries and preventing jackknifing during reverse maneuvers. Additionally, we show that by adjusting the weight $q_2^c$, the semitrailer’s position can be prioritized.

\begin{figure}[t]
    \centering
    \begin{subfigure}{0.32\columnwidth}
        \centering
        \includegraphics[width=\linewidth]{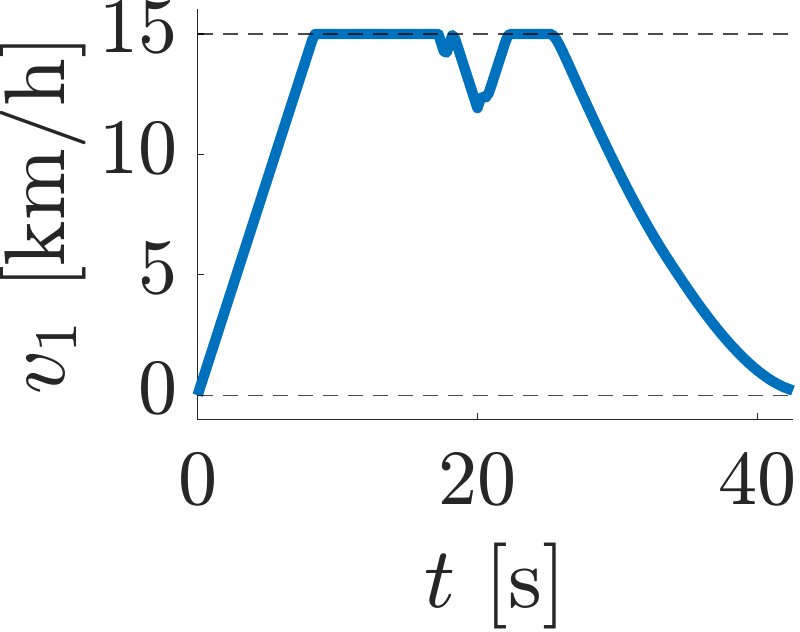}
    \end{subfigure}
    \hfill
    \begin{subfigure}{0.32\columnwidth}
        \centering
        \includegraphics[width=\linewidth]{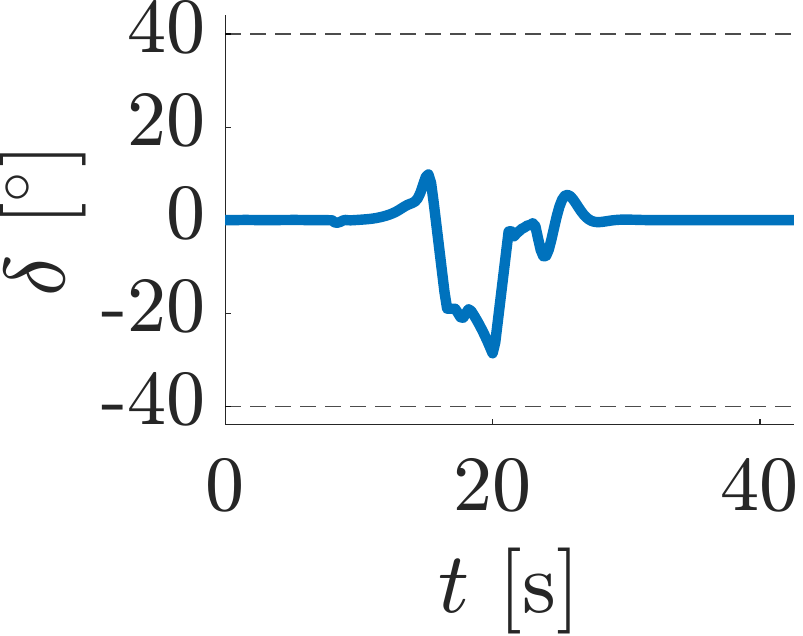}
    \end{subfigure}
    \hfill
    \begin{subfigure}{0.32\columnwidth}
        \centering
        \includegraphics[width=\linewidth]{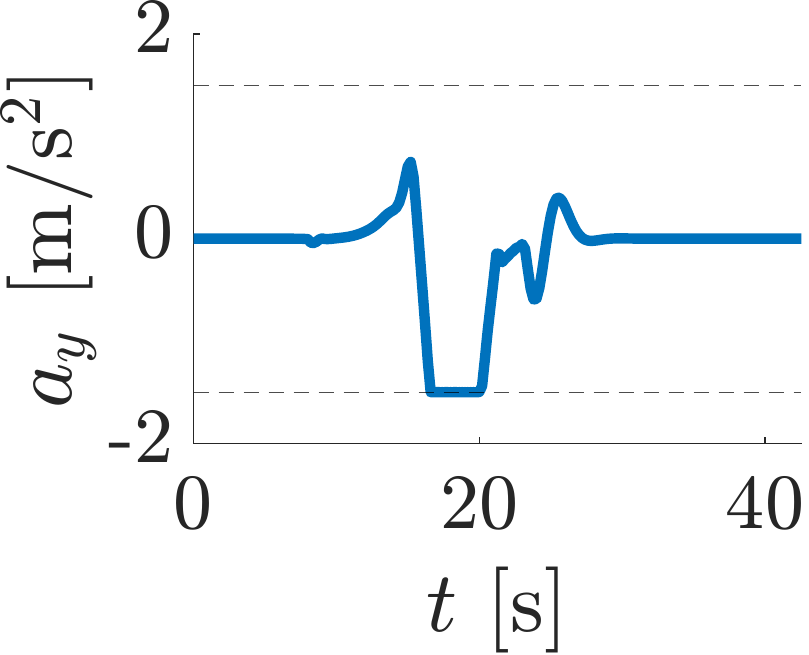}
    \end{subfigure}
    \caption{The velocity $v_1$, steering angle $\delta$, and lateral acceleration $a_y$ are plotted over time for the forward driving results shown in \Figure~\ref{fig:results_DD}. The horizontal dashed lines indicate the respective limit values. The figure shows that the lateral acceleration reaches its limit of 1.5m/s$^2$ between 16.5s and 20~s. To remain within this limit while achieving the steering angle required to take the turn, the velocity is decreased.}
    \label{fig:results_forward}
\end{figure}

 The vehicle successfully enters the premises in both forward and reverse directions, as shown in \Figure \ref{fig:results_DD}, using the same set of optimization weights. In both cases, the tractor adjusts its trajectory to keep the semitrailer axle within the corridor while negotiating the corner. During forward entry, the vehicle slows to 12~km/h to comply with the lateral acceleration limit, as shown in \Figure \ref{fig:results_forward}. The swept path visualizes the necessary maneuver space and highlights how the vehicle body extends beyond the corridor boundaries. This emphasizes the importance of accounting for vehicle size and kinematics when defining the corridor. Videos of the simulations are available at \texttt{https://gitlab.tue.nl/s169049/itsc2025}.

\begin{figure}[t]
    \centering
    \begin{subfigure}{0.48\columnwidth}
        \centering
        \includegraphics[width=0.95\linewidth]{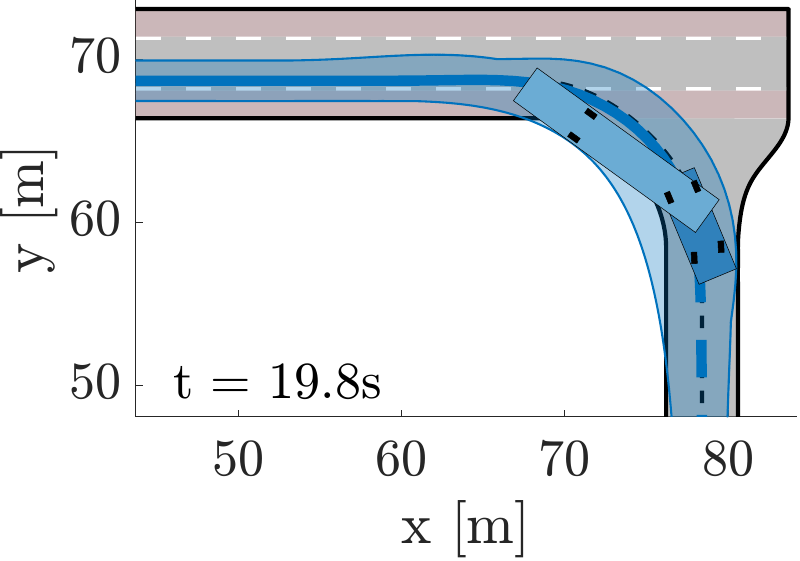}
    \end{subfigure}
    \hfill
    \begin{subfigure}{0.48\columnwidth}
        \centering
        \includegraphics[width=0.95\linewidth]{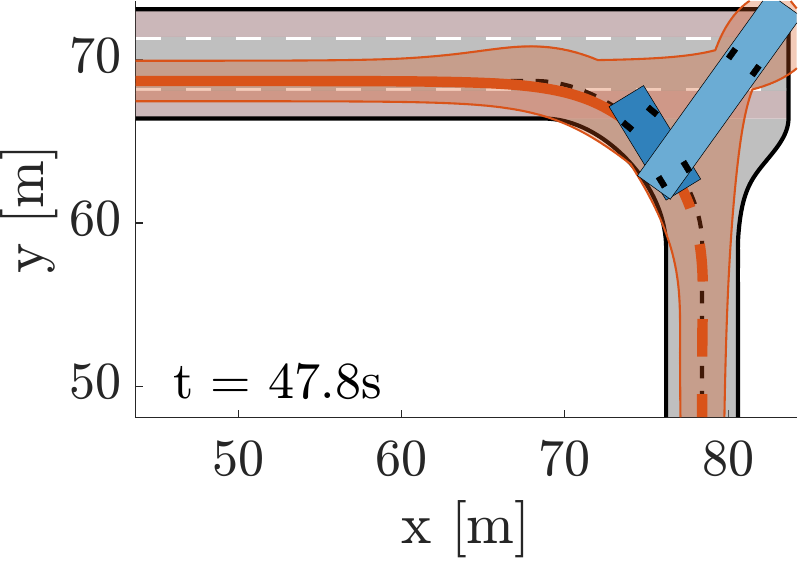}
    \end{subfigure}
    \caption{Simulation results for forward driving (\forward) and reverse driving (\reverse) when semitrailer is ignored in the optimization, resulting in corner cutting during forward driving and jackknifing in reverse driving.}
    \label{fig:results_failure}
\end{figure}

\Figure \ref{fig:results_failure} presents the forward and reverse driving results when the corridor constraints in \eqref{eq:corridor_cons} for $i=2$ are disabled and $q^l_2=q^c_2=0$. This means that the semitrailer axle is ignored during optimization, resulting in (self-)collisions as visualized in \Figure \ref{fig:results_failure}. In the forward maneuver, the semitrailer exits the corridor, while in the reverse maneuver, the vehicle jackknifes. These results highlight the importance of considering the semitrailer in trajectory planning to obtain safe trajectories.

\begin{figure}[t]
\centering
\begin{subfigure}{0.48\columnwidth}
    \centering
    \caption{}
    \includegraphics[width=0.95\textwidth]{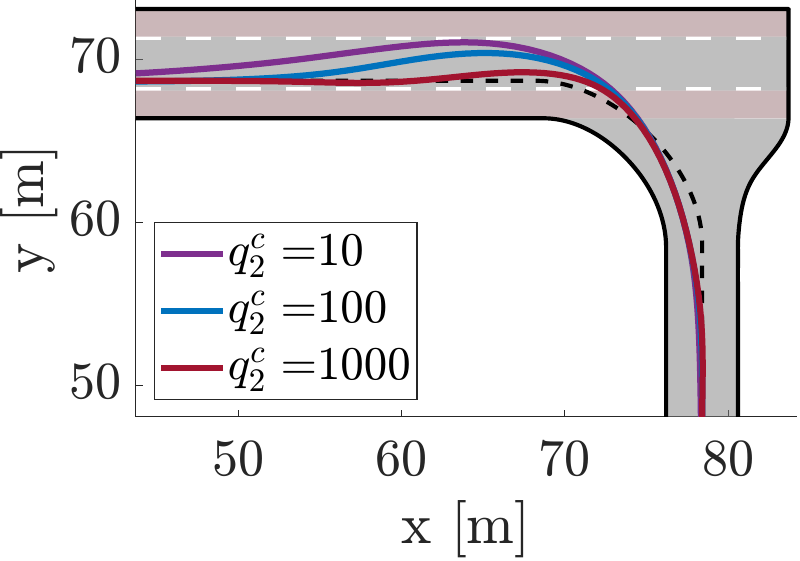}
    \label{fig:results_weighing1}
\end{subfigure}
\begin{subfigure}{0.48\columnwidth}
    \centering
    \caption{}
    \includegraphics[width=0.95\textwidth]{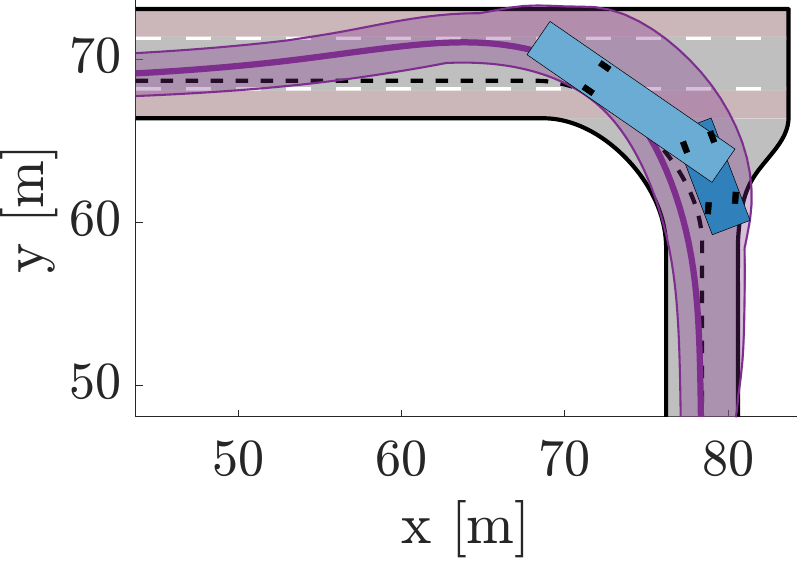}

    \label{fig:results_weighing2}
\end{subfigure}
\\
\begin{subfigure}{0.48\columnwidth}
    \centering
    \caption{}
    \includegraphics[width=0.95\textwidth]{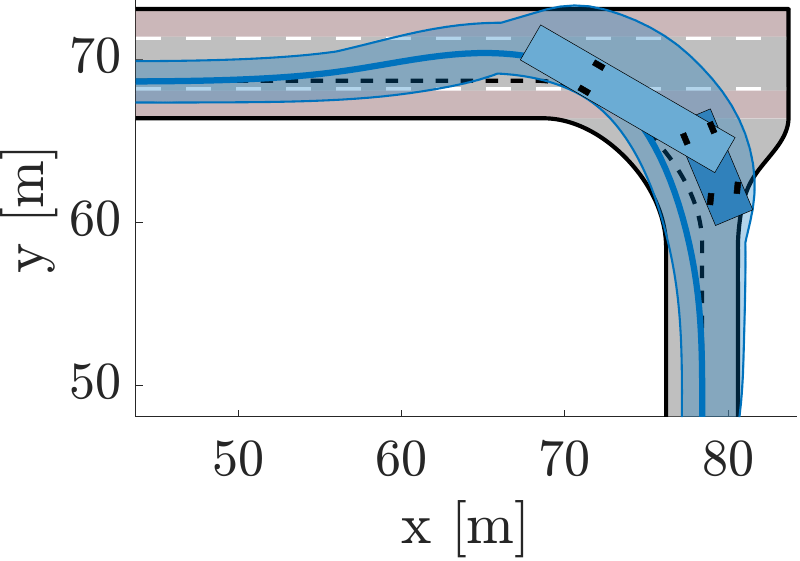}
    
    \label{fig:results_weighing3}
\end{subfigure}
\begin{subfigure}{0.48\columnwidth}
    \centering
    \caption{}
    \includegraphics[width=0.95\textwidth]{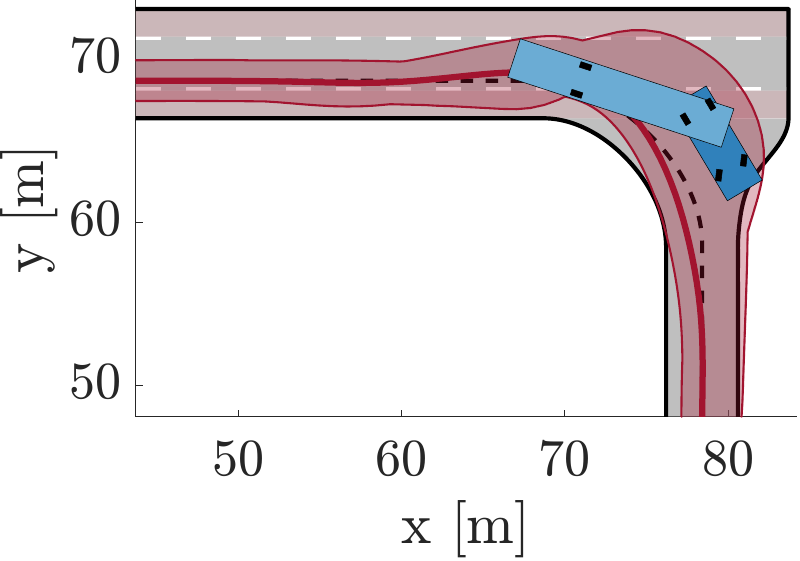}
    \label{fig:results_weighing4}
\end{subfigure}
\caption{Simulation results for forward driving with varying $q^c_2 \in \{10, 100, 1000\}$. The top-left plot shows semitrailer axle paths, while the other three depict the vehicle positions at $t=20$ s and complete swept paths.}
\label{fig:results_weighing}
\end{figure}

\begin{figure}[t]
    \centering
    \begin{subfigure}{0.32\columnwidth}
        \centering
        \includegraphics[width=\linewidth]{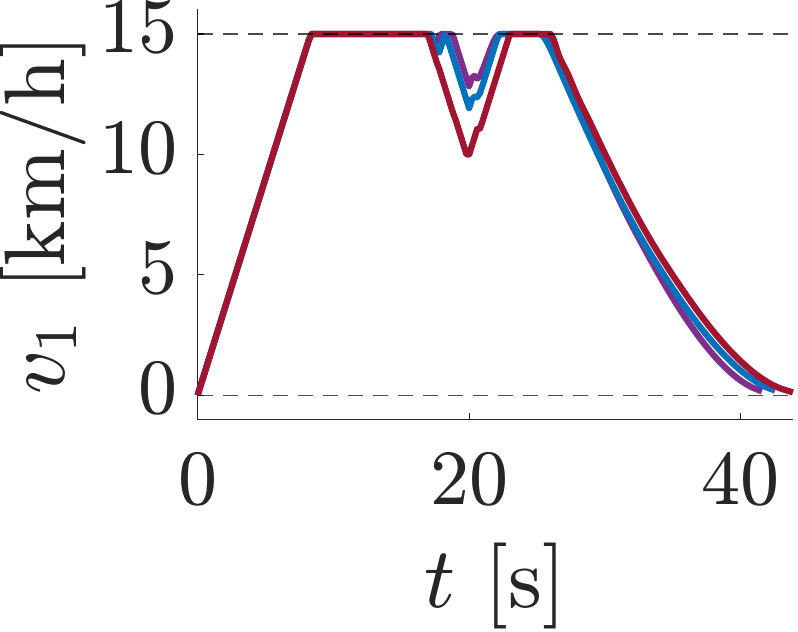}
    \end{subfigure}
    \hfill
    \begin{subfigure}{0.32\columnwidth}
        \centering
        \includegraphics[width=\linewidth]{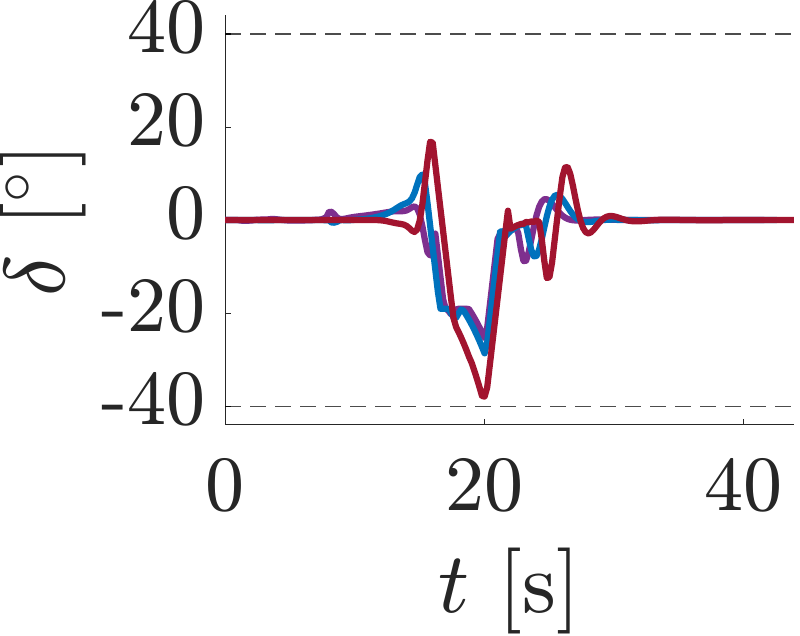}
    \end{subfigure}
    \hfill
    \begin{subfigure}{0.32\columnwidth}
        \centering
        \includegraphics[width=\linewidth]{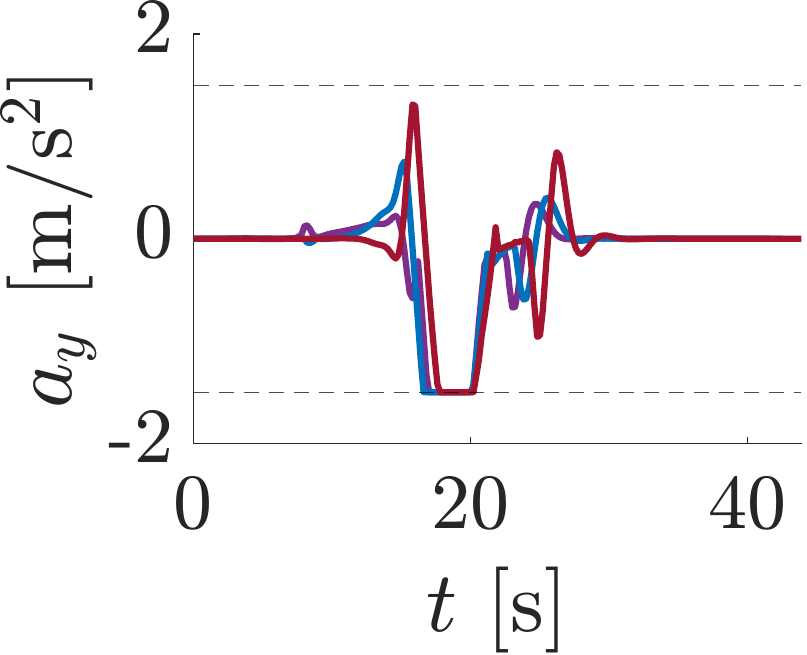}
    \end{subfigure}
    \caption{The velocity $v_1$, steering angle $\delta$, and lateral acceleration $a_y$ are plotted over time for the  results shown in \Figure~\ref{fig:results_weighing} for $q^c_2 = 10$ (\qcone), $q^c_2 = 100$ (\qctwo) and $q^c_2 = 1000$ (\qcthree). The horizontal dashed lines indicate the respective limit values.}
    \label{fig:results_weighing_v_delta_a}
\end{figure}

To demonstrate the effect of prioritizing a vehicle anchor point, we vary the weight of the semitrailer axle $q^c_2 \in\{10, 100, 1000\}$, while keeping all other parameters unchanged. As illustrated in \Figure \ref{fig:results_weighing1}, increasing $q^c_2$ reduces the maximum lateral deviation of the semitrailer axle from the reference path, as expected from \eqref{eq:error_i}. As a result, the tractor trajectory and the swept path are also influenced, as shown in \Figure \ref{fig:results_weighing2} to \ref{fig:results_weighing4}. Since a higher $q^c_2$ reduces the lateral deviation of the semitrailer, the tractor extends less into the opposite lane. As a result, the steering angle must increase to complete the turn, leading to a lower velocity in the corner to satisfy the lateral acceleration constraint, as visualized in \Figure \ref{fig:results_weighing_v_delta_a}. The semitrailer trajectories converge approximately halfway through the corner into the narrower section of the corridor for each $q^c_2$. This suggests that this specific corridor primarily governs the trajectory of the vehicle. Despite this, the influence of the variation in $q^c_2$ remains evident from the results. This shows that the proposed approach effectively changes the behavior of the vehicle by adjusting the individual weights and allows the prioritization of specific vehicle anchor points depending on the driving scenario. 

\begin{figure}[t]
    \centering
    \includegraphics[width=1.0\columnwidth]{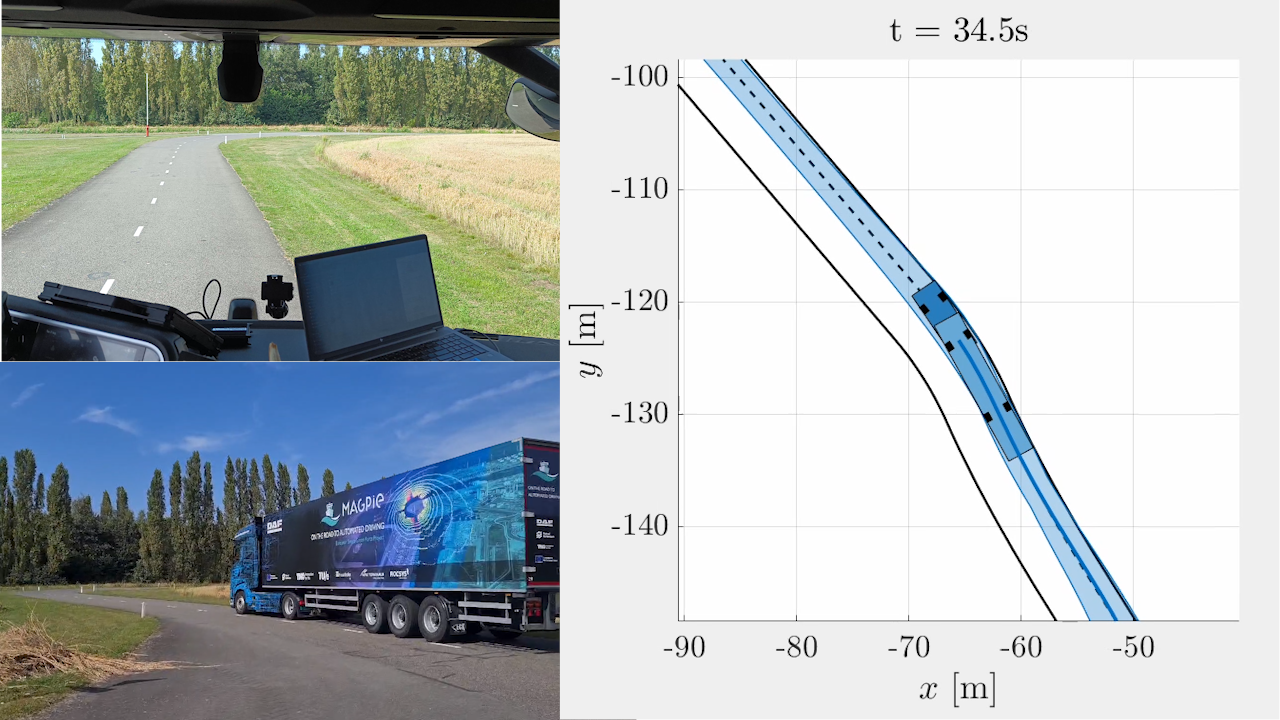}
    \caption{Still frame from the experimental testing video showing forward‑facing and side‑view camera feeds of the prototype vehicle negotiating a curve alongside a plot of its driven path (solid blue), reference path (dashed), swept path (shaded) and corridor boundaries (black)}
    \label{fig:experiment_result}
\end{figure}

Next to the simulations, the MPCC approach has been successfully implemented on a prototype tractor-semitrailer, which is visualized in \Figure \ref{fig:proto}. A video of the first implementation results can be found at \texttt{https://gitlab.tue.nl/s169049/itsc2025}, and a screenshot of this video is given in \Figure \ref{fig:experiment_result}. The planned trajectory is provided to the vehicle, where a controller is used to follow the trajectory. A new trajectory is planned once the vehicle has driven 10\% of the previous trajectory.

\section{CONCLUSIONS AND FUTURE WORK} \label{sec:conc}
This paper presented a trajectory planning method for articulated commercial vehicles using Model Predictive Contouring Control (MPCC). The main contribution is the integration of multiple progress states, one for each vehicle anchor point, into the optimization framework. This approach allows to ensure that the considered wheels remain within the drivable corridor and enables prioritization of vehicle anchor points based on driving scenarios. The simulation results show that the vehicle successfully navigates a given drivable corridor in both forward and reverse directions, highlighting the need to consider the semitrailer to avoid exiting the corridor or jackknifing. In addition, the influence of a single optimization weight on the trajectories is analyzed, providing insights into controlling vehicle behavior. Finally, the first full-scale prototype test demonstrates the practical applicability of the method.

In future work, our objective is to evaluate the proposed method in various driving scenarios through real-world testing, including dynamic environments with other (vulnerable) road users. Although constructing a drivable corridor, which incorporates vehicle size and oversway, from the local road model is central to our method, doing so robustly in all driving scenarios is not trivial and requires further investigation.

% Furthermore, generating a drivable corridor from the local road model, while accounting for size and overswaying of the vehicle, is central to the approach but might not be trivial to create in various driving scenarios. For example,  

% present challenges that necessitate further investigation. \textcolor{red}{For example, when negotiating a 90\degree~turn in a narrow environment, the vehicle may either cut the corner or enter into the opposite lane. Cutting the corner can be acceptable if it drives onto grass or dirt, but undesirable if it drives onto a sidewalk, although in some cases that may be the only option.}

\addtolength{\textheight}{0cm}   % This command serves to balance the column lengths
                                  % on the last page of the document manually. It shortens
                                  % the textheight of the last page by a suitable amount.
                                  % This command does not take effect until the next page
                                  % so it should come on the page before the last. Make
                                  % sure that you do not shorten the textheight too much.

%\section*{ACKNOWLEDGMENT}
%MODI project + DAF S\&DC group

%%%%%%%%%%%%%%%%%%%%%%%%%%%%%%%%%%%%%%%%%%%%%%%%%%%%%%%%%%%%%%%%%%%%%%%%%%%%%%%%
\bibliographystyle{plain} % We choose the "plain" reference style
\bibliography{refs} % Entries are in the refs.bib file

\end{document}